# Deep Collaborative Multi-modal Learning for Unsupervised Kinship Estimation

Guan-Nan Dong, Chi-Man Pun, *Senior Member, IEEE*, Zheng Zhang, *Senior Member, IEEE*

*Abstract*—Kinship verification is a long-standing research challenge in computer vision. The visual differences presented to the face have a significant effect on the recognition capabilities of the kinship systems. We argue that aggregating multiple visual knowledge can better describe the characteristics of the subject for precise kinship identification. Typically, the age-invariant features can represent more natural facial details. Such age-related transformations are essential for face recognition due to the biological effects of aging. However, the existing methods mainly focus on employing the single-view image features for kinship identification, while more meaningful visual properties such as race and age are directly ignored in the feature learning step. To this end, we propose a novel deep collaborative multi-modal learning (DCML) to integrate the underlying information presented in facial properties in an adaptive manner to strengthen the facial details for effective unsupervised kinship verification. Specifically, we construct a well-designed adaptive feature fusion mechanism, which can jointly leverage the complementary properties from different visual perspectives to produce composite features and draw greater attention to the most informative components of spatial feature maps. Particularly, an adaptive weighting strategy is developed based on a novel attention mechanism, which can enhance the dependencies between different properties by decreasing the information redundancy in channels in a self-adaptive manner. Moreover, we propose to use self-supervised learning to further explore the intrinsic semantics embedded in raw data and enrich the diversity of samples. As such, we could further improve the representation capabilities of kinship feature learning and mitigate the multiple variations from original visual images. To validate the effectiveness of the proposed method, extensive experimental evaluations conducted on four widely-used datasets show that our DCML method is always superior to some state-of-the-art kinship verification methods.

*Index Terms*—kinship verification, information security, self-supervised learning

## I. INTRODUCTION

RECENT years has witnessed the emerging prosperity of kinship verification, and extensive efforts have been devoted to improving the robustness of the kinship feature learning and relationship estimation. Typically, kinship verification is to validate if two persons are biologically related by measuring their similarity. Due to the immense amount of social cases presence such as missing child search, social media information, family photo annotation, child recovery, and criminal trial, recognizing kin relations [1], [2] has attracted tremendous research interest from both academia and industry communities. Numerous algorithms of kinship verification have been proposed in the past few years, including kinship verification from a single feature extractor to multiple feature extractors. Compared to the single feature extractor, multiple feature extractors are more challenging and feasible in real applications. Although the difficulty of the kinship algorithm is to excavate the hidden similarity inherited shown in the different faces with a large inter-class appearance variance, the existing methods have still achieved encouraging performance. According to the difference of feature extractor, a general recognition framework appeared in most of the previous works can be categorized as follows: 1) shallow learning models [3], [4], and 2) deep learning models [5], [6].

The shallow learning models mainly aim at extracting discriminative features by handcrafted descriptors, such as LBP [7] and HOG [8]. For example, the spatial pyramid learning (SPLE) [3] integrates spatial learning and pyramid learning to extract multiple features for kinship verification. The discriminative multi-metric learning (DMML) [4] applies different feature extractors (such as LBP and HOG) to maximize the compactness of the intra-class distance and the separability of the inter-class distance. Although these models also realize appealing performance, these models have a poor generalization ability due to the fundamental low-level features. In other words, these methods only could be adopted to overcoming specific issues since the features need to be manually designed and rely on prior knowledge. Moreover, handcrafted features need to trade-off the correlation between effectiveness and robustness. Hence, they lack the flexibility for the more complicated computer vision tasks.

By contrast, the deep learning paradigm can create feasible networks for various practical computer vision tasks due to the powerful representation capabilities. Moreover, deep learning models can capture the high-level features from observations, which have more robustness to intra-class variability. Notwithstanding, deep learning models can transfer knowledge to other datasets and tasks for improving their generalization ability. For example, an end-to-end CNN-Basic [5] is employed to capture high-level and more detailed features under the guidance of loss function to maximize the intra-class coherence. The deep kinship verification (DKV) [6] uses a separated learning strategy, i.e., an auto-encoder network used for feature extraction and deep metric learning.

Although extensive studies have been devoted to improving the robustness and discriminant of kinship verification systems, the existing works are mainly operated on single-modality feature learning, which fails to fully explore the underlying characteristics of kinship facial images and leads to inferior kinship estimation results. In particular, multi-modal

Guan-Nan Dong, Chi-Man Pun, and Zheng Zhang are with the Department of Computer and Information Science, University of Macau, Macau 999078, China. (e-mail: guannandong@outlook.com, cmpun@umac.mo, darrenzz219@gmail.com).

learning [9], [10] has achieved excellent performance in object classification. As such, it is urgent to explore multi-modal collaborative learning for effective kinship verification. Moreover, for multi-modal learning, how to effectively integrate multiple modalities into one unified learning space is still an open research problem. Most existing works tend to employ the concatenation or manually-assigned weighting strategies on multiple modalities to produce composite features. However, these fusion schemes are hard to fully exploit the complementary information and the correlations provided by multiple modalities and may increase the information redundancy between channels. Additionally, kinship estimation is always based on supervised learning and relies on a large quantum of labeled pairwise data for model training. By contrast, the existing supervised learning features have not been sufficient to completely represent all the facial details and defense the large variations of face pictures, such as rotation, mask, and expressions. Hence, it is still a challenge on how to capture and generate more valid semantic features to further improve the kinship validation performance.

To overcome the above deficiencies, in this paper, we propose a novel deep collaborative multi-modal learning (DCML) method to promote the generation of more enriched features. In the method, we leverage the complementary and correlations of the multiple modalities to aggregate multiple visual information in a self-adaptive learning manner. Specifically, we introduce a collaborative multi-modal learning strategy, which can ensure the semantic and feature consistency of different modalities to capture their complementary traits. Meanwhile, the abundant correlation knowledge across different modalities are well preserved in the shared learned features. Moreover, we propose an adaptive feature fusion mechanism to weight the importance of each visual feature, in which an adaptive weighting strategy is designed to enhance discriminative information and filter out contradictory components. In this way, the network can flexibly perceive the dependencies between features to promote the distinguishable ability of the learned features. Furthermore, we employ the self-supervised learning method to take full advantage of internal structures of data instead of data itself, which can alleviate the overfitting problem and disturbance problem. Notably, the self-supervised learning method can generate more sample pairs to eliminate the limitation of insufficient data. The outline of the proposed method is graphically illustrated in Figure 1. The experimental results show the feasibility and superiority of our DCML compared to some state-of-the-art kinship verification methods.

The main contributions of this paper are summarized as follows:
- We propose a novel deep collaborative multi-modal learning (DCML) method for effective unsupervised kinship verification. We incorporate multi-modal features with self-adaptive learning in self-supervised learning, which can attract more attention to the most informative components across different modalities and strengthen the representation capabilities of the learned features. To the best our knowledge, *this is the very first attempt* that leverages the multi-modal information and self-supervised learning technique to guide kinship verification.
- We develop a novel adaptive weighting strategy to handle the multi-modal information reasonably, which can flexibly evaluate the importance of multi-modal features for collaborative feature learning in a self-adaptive manner.
- To overcome the deficiencies of supervised learning and improve the robustness of the learning model on image disturbance, we employ the self-supervised learning method to explore the deeper internal structure of data by finding the correlations between samples. At the same time, it can generate self-learned semantic information from raw data to promote the representation capabilities of features.
- Comprehensive experimental evaluations are conducted on four widely-used datasets. The qualitative and quantitative experiments demonstrate the effectiveness and superiority of our DCML method over some state-of-the-art kinship verification methods.

The remainder of this paper is organized as follows: Section II introduces related work in recent years. Section III explicitly presents our proposed method. Section IV validates the performance of our method by extensive experiments and gives the experimental analysis. Section V concludes our paper in research, and we also discuss the challenges in this area and scope for further study.

## II. RELATED WORK

In the last decades, many attentions have been applied to kinship verification under a number of learning structures. This section briefly reviews two related research topics: 1) kinship verification, and 2) self-supervised learning.

### A. Kinship Verification

In the human face analysis, people who belong to the same family have similar familial traits, especially their facial details. From this inspiration, many related researches have revealed that kinship analysis broadens the knowledge of facial structure. Human faces similarity is a potential cue to verify whether they are biologically related or not. Due to the frequent changes of aging and the manner of taking and illumination, kinship verification meets various challenges. Some seminal attempts have been developed to develop an effective and realizable method for kinship verification. Existing methods can be categorized as follows: 1) shallow learning methods, and 2) deep learning methods.

Generally, shallow learning methods generally use the traditional descriptors such as LBP (local binary pattern) [1], SIFT (scale-invariant feature transform) [11] [1] [12], HOG (histogram of gradient) [11] [1] to extract shallow handcrafted features. Popular algorithms such as [1], [3], [4], [11]–[16] have been attempted to explore the kinship relations. Technically, the shallow learning methods focus on extracting linear and pre-defined features. Some nonlinear variations and most significant traits can not be expressed very well. Therefore, traditional descriptors can not capture sufficient representation abilities, especially in large-scale applications. Notably, compared to shallow algorithms depending on prior knowledge, deep-learning models are better when excavating the hidden similarity inherited between parent-offspring facial images, and they can represent the features in a higher-level technology. Typical algorithms such as [5], [6], [10], [17]–[27] have been developed to exploit more complementary information rather than staying on the original low-level features. For example, an end-to-end manner [5] is the first attempt to treat the kinship problem. It uses a unified framework to cope with kinship verification by way of binary classification. The manner of end-to-end learning is convenient, and manual human intervention is unnecessary. However, most similarity verification is seldom trained in this way. More current works incline to employ the separated learning strategy to express the common features between samples. For example, the proposed method [25] employs CNN as a feature extractor and takes NRML [11] as a similarity measurement to process features in a jointly learning manner.

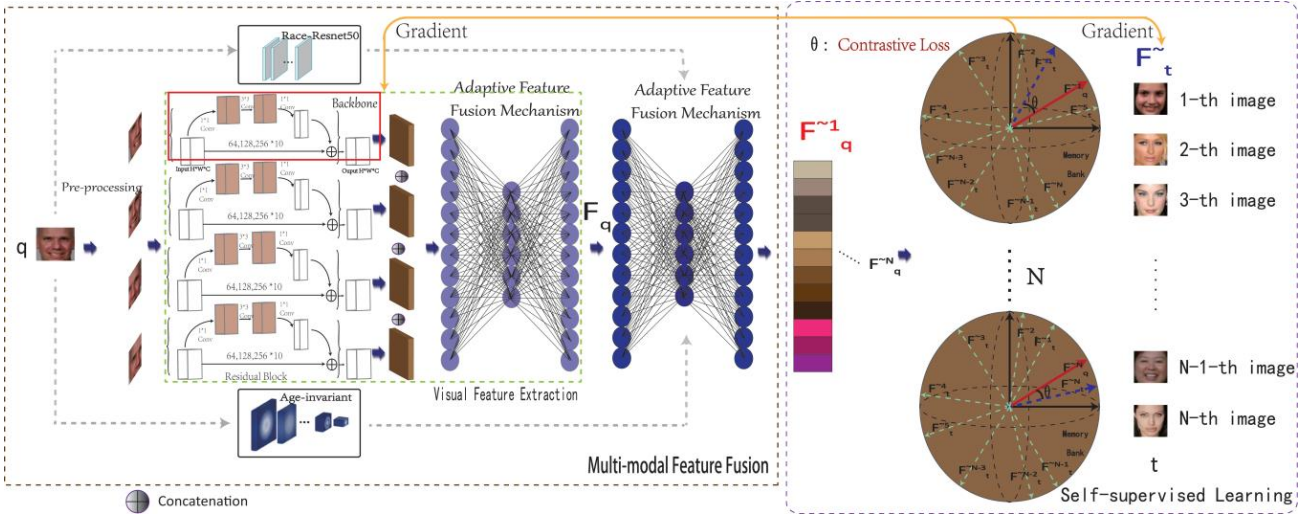

Fig. 1: The overview of our proposed framework. An end-to-end method includes multi-modal feature collaborative learning and self-supervised learning to guide kinship verification.

Generally, these methods can excavate the interior structure between two facial images and improve the distinguishable ability of the learned features. However, these methods incline to extract the single visual features but fail to explore the multiple modalities features such as facial attributes, which leads to unsatisfied performance for some complicated kinship tasks. Besides, most of the existing works tend to employ the concatenation or manually-assigned weighting strategies to preserve abundant correlation knowledge of samples from different perspectives. Nevertheless, these fusion schemes can not fully reflect the intrinsic correlations between multiple features and can not filter out the information redundancy between channels. Hence, these methods are hard to capture complementary information, and internal correlations between multiple modalities are also under-explored, leading to inferior performance.

*B. Self-supervised Learning*

In the past decades, many researchers have achieved outstanding results in complicated computer vision tasks by performing supervised learning. These visual tasks need a large quantum of labeled data to train and improve learning. However, getting a valid dataset is immensely time-consuming and expensive, and the annotations are also laborious and tedious. For example, ImageNet [28] includes $14M$ images and label them taking roughly 22 human years. It is notable that a decent amount of tasks do not have enough data and are not easy to collect labels. Hence, it is still a challenge to obtain abundant data and labels. To mitigate the above deficiencies, unsupervised learning is proposed to exploit the nature of the interior structure of categories to train a model instead of relying on some complementary information provided by a large dataset. By contrast, the performance of unsupervised learning is far from satisfactory, and they are much less efficient due to no guidance of the semantic labels. Therefore, some researchers argue that these tasks can leverage supervised learning to train unsupervised dataset. In this way, we will have all the information, including data and pseudo-labels. This idea has been intensively studied in many computer vision tasks, called self-supervised learning, but *has not* successfully applied in the difficult kinship estimation task.

The self-supervised learning has been introduced to learn visual tasks and made great achievements in image processing. Specifically, to learn informative image representations, such tasks can be categorized as follows: 1) Pretext Task: Predicting Rotation of Images [29], Colourisation [30], Context Prediction [31] and Jigsaw [32]; 2) Contrastive Learning: Contrastive Predictive Coding (CPC) [33], Non-Parametric Instance Discrimination [34], Momentum Contrast (MoCo) [35] and SimCLR [36]. For video representation learning, the tasks generally lie in the following situations: 1) Object tracking [37], [38]; 2) Ego-motion [39], [40]. More other related works could be found in a recent survey paper [41].

Kinship verification could be considered as a few-shot classification problem. Therefore, benefiting from the advantages of self-supervised learning, we, *for the first time*, employ such an effective learning to enrich the diversity of samples and fully exploit the internal structure of the dataset to learn high-quality semantics. In sharp contrast to previous kinship works based on supervised learning, we use unsupervised learning to obtain powerful representation capabilities to promote the distinguishable ability of the learned features and alleviate the effect of latent variations in feature learning.

## III. PROPOSED METHOD

### A. Overview of the Proposed Method

To obtain more complementary information and face cues, we propose a novel unsupervised learning method called deep collaborative multi-modal learning (DCML) to enhance the information representation by aggregating multiple modalities knowledge for kinship verification. Specifically, we have three modality extraction modules *i.e.*, race ResNet-50 module, Age-invariant module, and facial visual feature extraction module to capture the race features, the de-aging features, and the original facial visual features based on the patch. Moreover, we introduce an effective adaptive feature fusion mechanism by performing an adaptive weighting strategy to get refined facial visual composite features and unified multi-modal feature representations. Furthermore, to alleviate the problem of over-fitting, we employ the outstanding self-supervised learning diagram to enrich the diversity of samples to guide and enhance the discriminative ability of learned features. The outline of the proposed method is graphically illustrated in Figure 1. Figure 2 shows the proposed adaptive feature fusion mechanism.

### B. Multi-modal Feature Extraction

Kinship verification processing is not monotonous. There is a difference in the verification results because of the facial changes, especially in shape and face texture in childhood and old age. Hence, we use different modalities to represent visual information in a collaborative learning manner. This section introduces the feature learning steps of the following three parts: 1) facial image visual feature extraction, 2) de-aging feature extraction, and 3) race feature extraction.

*1) Facial image visual feature extraction:* Technically, the global receptive field is inconsistent with the local receptive field. The patch learning can limit the receptive field to specific areas so that the network can pay attention to the unique information from different patches. Compared with the information loss caused by global learning, patch learning can describe inconsistent information more precisely.

We define a human face dataset as $\mathbf{X} = \{\mathbf{x}^i \mid i = 1, 2, 3, \ldots, N\}$, where $\mathbf{x}^i \in \mathbb{R}^{H \times W \times C}$ is the $i$th sample. After cropped it to four over-lapping patches, we have $\mathbf{X} = \{\mathbf{x}_n^i \mid n = 1, 2, 3, 4 \ \& \ i = 1, 2, 3, \ldots, N\}$, where $\mathbf{x}_n^i \in \mathbb{R}^{h \times w \times C}$. As shown in Figure 1, we use a generic framework with four branches to extract patch features. As shown in Table III, each backbone is a 30-block ResNet structure. It consists of 3 stages with respectively 10, 10, 10 stacked residual blocks. Each residual block has 3 Conv units respectively 1*1, 3*3, 1*1. Finally, a linear layer outputs the initial face features with 256 dimension.

Given an image patch $\mathbf{x}_n^i \in \mathbb{R}^{h \times w \times C}$ as an input for the network, which can be mapped from $\mathbf{x}_n^i \in \mathbb{R}^{h \times w \times C}$ to $\mathbf{U}_n^i = \mathbf{f}(\mathbf{x}_n^i)$ by a transformation $\mathbf{f}(\cdot)$, where $\mathbf{U}_n^i \in \mathbb{R}^{1 \times 1 \times c}$; $\mathbf{f}(\cdot) = \Theta^\top \mathbf{X} + \mathbf{b}$, $\Theta$ is a set of the weights; $\mathbf{b}$ is a set of the bias. The original visual features of the $i$th samples can be written as:

$$\mathsf{F}^i = \langle \mathbf{U}_1^i, \mathbf{U}_2^i, \mathbf{U}_3^i, \mathbf{U}_4^i \rangle \quad (1)$$

where $\langle \cdot \rangle$ is a concatenation operation, and $\mathsf{F}^i \in \mathbb{R}^{1 \times 1 \times 4c}$. To make better use of the aggregated information, we introduce an adaptive feature fusion mechanism to furthest capture channel-wise dependencies between different patches. The details are illustrated in Figure 2. The adaptive feature fusion mechanism can be shown as:

$$\begin{aligned} \mathbf{z}^i &= \Phi(\mathsf{F}^i) \\ \mathbf{s}^i &= \Psi(\mathbf{z}^i, \Theta^i) = \sigma(\delta_2(\delta_1(\mathbf{z}^i))) \\ \mathbf{F}^i &= \mathbf{s}^i * \mathsf{F}^i \end{aligned} \quad (2)$$

where $*$ denotes channel-wise multiplication between scalar $\mathbf{s}^i$ and the feature vector $\mathsf{F}^i$; $\mathbf{z}^i, \mathbf{s}^i \in \mathbb{R}^{4c}$. $\Phi(\cdot)$ is a channel descriptor which can generate channel-wise statistics by a global average pooling layer. $\Psi$ is a nonlinear interaction learning function achieved by performing two fully-connected (FC) layers $\delta_1, \delta_2$ with a sigmoid activation. $\delta_1$ is used to reduce the dimensionality with ratio $r_1$ (here, $\mathbf{z}^i \in \mathbb{R}^{4c \to \frac{4c}{r_1}}$), and $\delta_2$ is designed to increase the dimensionality with ratio $r_1$ (here, $\mathbf{z}^i \in \mathbb{R}^{\frac{4c}{r_1} \to 4c}$), and this ratio choice will be discussed in the Experiments section. $\mathbf{F}^i \in \mathbb{R}^{1 \times 1 \times d_1}$ is the final output of feature fusion by the proposed adaptive feature fusion mechanism.

*2) De-aging Feature Extraction:* Kinship verification intrinsically has more severe intra-class variations and smaller inter-class variations than the general face verification problem. To overcome this challenge, we use a multi-modal feature integration strategy to enhance the feature representation for kinship verification. Specifically, due to the biological effects of aging, we leverage the de-aging features to capture the real facial invariant features.

For de-aging feature extraction, inspired by Decorrelated Adversarial Learning (DAL) [42], they argue that the composition of age is linear. It can be factorized into age-dependent component and age-invariant component. The age-dependent component describes the age variations, and the age-invariant component describes identity information. The age-invariant component is our need. We use the proposed method to extract age-invariant features. Given an input image $\mathbf{x}^i \in \mathbb{R}^{H \times W \times C}$, we feed it to the backbone $\mathbf{K}$ as in DAL [42]: $\mathbf{f}_f^i = \mathbf{K}(\mathbf{x}^i)$, where $\mathbf{f}_f^i \in \mathbb{R}^{d_2}, \forall f \in \{id, age\}$, $\mathbf{f}_{id}^i$ represents the age-invariant component, and $\mathbf{f}_{age}^i$ denotes the age-dependent component, *i.e.*,

$$\begin{cases} \mathbf{f}_{age}^i = \mathsf{R}(\mathbf{f}_f^i) \\ \mathbf{f}_{id}^i = \mathbf{f}_f^i - \mathsf{R}(\mathbf{f}_f^i), \end{cases} \quad (3)$$

where $\mathsf{R}$ is the Residual Factorization Module with two stacked FC-ReLU layers, which can be performed to obtain different age-related components. According to the Decorrelated Adversarial Learning (DAL) [42], we get the final age-invariant component $\mathbf{f}_{id}^i$ by using

$$\min_{K,R} \max_{C} (|\rho(\mathsf{C}(\mathbf{f}_{id}^i)), \mathsf{C}(\mathbf{f}_{age}^i)|),$$
$$\rho = \frac{\sqrt{\mathbf{Cov}(\mathsf{C}(\mathbf{f}_{id}^i), \mathsf{C}(\mathbf{f}_{age}^i))}}{\mathbf{Var}(\mathsf{C}(\mathbf{f}_{id}^i))\mathbf{Var}(\mathsf{C}(\mathbf{f}_{age}^i))}, \quad (4)$$

where $\mathsf{C}$ is a canonical mapping module with three stacked FC-ReLU layers, and $\rho$ is a canonical correlation. Considering the composition of age is linear, $\mathbf{f}_{id}^i$ and $\mathbf{f}_{age}^i$ have latent relationship with each other. However, the age-invariant component should be invariable when the age-dependent component is changing. The two components should be mutually uncorrelated. Hence, canonical mapping is used to find the maximum of $\rho$ to makes the correlation between $\mathbf{f}_{id}$ and $\mathbf{f}_{age}$ minimum.

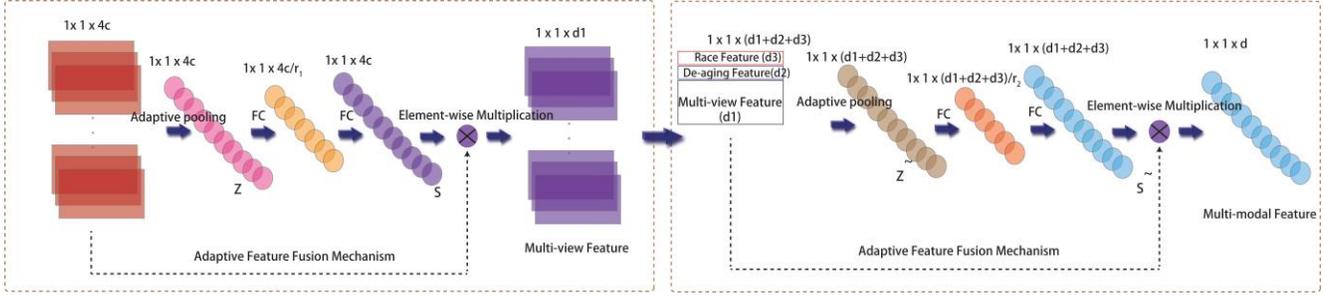

Fig. 2: The proposed adaptive feature fusion mechanism for the adaptively-weighted feature fusion.

*3) Race Feature Extraction:* For the race feature extraction, we employ the well-known **Resnet-50** [43] as our backbone network. The network is pre-trained on **ImageNet** [28] dataset with freezed weights and bias first and removes the last fully-connected layer by replacing it with a global avgpooling layer instead of flattening directly. Then, we use **CACD** [44] as our basic race datasets to train Resnet-50. We manually divide the **CACD** [44] into three categories: Asian, African and Caucasian for race classification. Because this dataset has 16M images, it is sufficient for us to train race classification models. Given an input image $\mathbf{x}^i \in \mathbb{R}^{H \times W \times C}$, we feed it to the backbone $\mathsf{G}$, and we have

$$\mathbf{f}_{race}^i = \mathsf{G}(\mathbf{x}^i), \quad (5)$$

where $\mathbf{f}_{race}^i \in \mathbb{R}^{d_3}$. The final race features are formulated by $\mathbf{f}_{race}^i$.

### C. Adaptive Multi-modal Feature Fusion

The proposed adaptive feature fusion mechanism can take fully advantage of their implicit complementary characteristics to learn the significant details between various modalities. Such multi-modal feature fusion can effectively integrate multiple modalities into one unified learning space to share rich semantics and enhance the effectiveness of verification performance. The main structure of adaptive feature fusion mechanism is shown in Table I.

Based on the above multi-modal feature extraction, as shown in Figure 2, we can obtain three kinds of features:

TABLE I: Adaptive Multi-modal Feature Fusion.

| Output size | Layer Structure |
|---|---|
| $d_1 + d_2 + d_3$ | global average pool |
| $\frac{(d_1+d_2+d_3)}{r_2}$ | fc layer with decreasing reduction ratio $r_2$ |
| $d_1 + d_2 + d_3$ | fc layer with increasing reduction ratio $r_2$ |
| d | sigmod function |

$\mathbf{f}_{race}^i$, $\mathbf{F}^i$, and $\mathbf{f}_{id}^i$. According to Eq. (2), we have the original feature fusion as follows:

$$\mathsf{F}^i = \langle \mathbf{F}^i, \mathbf{f}_{id}^i, \mathbf{f}_{race}^i \rangle, \quad (6)$$

where $\langle \cdot \rangle$ is a concatenation operation. After the adaptive feature fusion mechanism, the adaptive weighting feature fusion can be formulated as follows:

$$\begin{aligned} \mathbf{z}^i &= \Phi(\mathsf{F}^i) \\ \mathbf{s}^i &= \Psi(\mathbf{z}^i, \Theta^i) = \sigma(\delta_2(\delta_1(\mathbf{z}^i))) \\ \mathbf{F}^i &= \mathbf{s}^i * \mathsf{F}^i \end{aligned} \quad (7)$$

where $*$ denotes the channel-wise multiplication between scalar $\mathbf{s}^i$ and the feature vector $\mathsf{F}^i$; $\mathbf{z}^i, \mathbf{s}^i \in \mathbb{R}^{d_1+d_2+d_3}$. $\Phi(\cdot)$ is a channel descriptor to generate channel-wise statistics. $\Psi$ is a nonlinear interaction learning function consisting of two fully-connected (FC) layers $\delta_1$, $\delta_2$ and a sigmoid activation. $\mathbf{F}^i \in \mathbb{R}^d$ is the final output of feature fusion. $\delta_1$ is used to reduce the dimensionality with a ratio $r_2$ (here, $\mathbf{z}^i \in \mathbb{R}^{d_1+d_2+d_3 \to \frac{d_1+d_2+d_3}{r_2}}$), $\delta_2$ is to increase the dimension with ratio $r_2$ (here, $\mathbf{z}^i \in \mathbb{R}^{\frac{d_1+d_2+d_3}{r_2} \to d_1+d_2+d_3}$), (this ratio choice will discussed in Section Experiments).

### D. Self-supervised Learning

We consider a batch-size sample pairs $\mathbf{Q} = \{\mathbf{q}^1, \mathbf{q}^2, \cdots, \mathbf{q}^N\}$ and $\mathbf{T} = \{\mathbf{t}^1, \mathbf{t}^2, \cdots, \mathbf{t}^N\}$ and treat $\mathbf{T}$ as a matching queue. Each sample will be fed into the backbone to learn their features. The matching queue will be progressively replaced when the batch-size samples are changing. Each $\mathbf{q}$ has $\mathbf{T}$ matching samples. Assume that $\mathbf{q}^i$ has a single positive match $\mathbf{t}^j$, where $i = j$. To improve the feature representation ability, we need a large memory bank to cover a rich set of generated sample pairs, which can promote the model dynamic evolution. We assume that each sample pair is a distinct class of its own. For batch-size $N$ samples, we can cover this batch-size images and have $N^2$ labels. Hence, for each $\mathbf{q}$, after fed into the framework ($\mathbf{F}_\mathbf{q}^i = f_\mathbf{q}(\mathbf{q})$, $f_\mathbf{q}$ is our DCML network), we have one positive sample pair and $(N-1)$ negative sample pairs, which are dissimilar to the $\mathbf{q}$ and all $\mathbf{T}$ ($\mathbf{F}_\mathbf{t}^i = f(\mathbf{t})$, $f$ is our DCML network) are necessary. The memory bank refers to **Moco** [35], which can leverage the momentum update to alleviate the inconsistency between outdated matching queues and the newest ones caused by different extractors all over the past epoch in the memory bank. The feature representation and model parameters will be updated by stochastic gradient descend during each iteration, and the memory bank will be updated with each batch-size samples. Each batch-size feature is a unit of updated vectors. We use a memory bank to store all sample pairs for contrastive learning, which can be generalized to new classes. As such, this method can make our attention entirely focus on the positive sample pair feature representation. Moreover, we minimize a noise-contrastive estimation, similar to InfoNCE [33], to guide our model. It can be defined as follows:

$$L_{NCE} = -\log \frac{\exp(\mathbf{F}_\mathbf{q}^i \cdot \mathbf{F}_\mathbf{t}^j / \tau)}{\sum_{j=0}^{N} \exp(\mathbf{F}_\mathbf{q}^i \cdot \mathbf{F}_\mathbf{t}^j / \tau)} \quad (8)$$

where $\tau = 0.07$ is a temperature hyper-parameter [45] to tune the concentration distribution of $\mathbf{t}^j$. To classify the sample pairs $\mathbf{Q}$ and $\mathbf{T}$, we leverage **Cosine Similarity** to measure similarity. Here, $\mathbf{F}_\mathbf{q}^i \cdot \mathbf{F}_\mathbf{t}^j \to \cos(\mathbf{F}_\mathbf{q}^i, \mathbf{F}_\mathbf{t}^j)$. When $i \neq j$, it demonstrates that the current $\mathbf{q}$ is not similar to the current $\mathbf{t}$, the value of $\exp(\mathbf{F}_\mathbf{q}^i \cdot \mathbf{F}_\mathbf{t}^j / \tau)$ should be regularized to the best of the minimum values. When $i = j$, it demonstrates that the current $\mathbf{q}$ is similar to the current $\mathbf{t}$, which makes $\exp(\mathbf{F}_\mathbf{q}^i \cdot \mathbf{F}_\mathbf{t}^j / \tau)$ be the best of the maximum values. Here log will be close to **1**, which minimizes the loss function. Moreover, we return a matrix with **N*N**, where each row corresponds to one sample, and we regulate the matrix which makes the **0**th column of each row be the true value. Precisely, the **1**st to **N-1**th columns are the negative samples. Notably, for this loss function, the true label of each sample pair is the **0**th column because the label is an **N*1** vector with all zeros.

*E. Loss Function*

To optimize multi-view feature selection and better describe the characteristics of the subject for precise kinship identification, we use the variable-controlling method to operate the loss functions of race and de-aging individually, which makes the race and de-aging features have invariability when training the image patches under self-supervised learning.

*1) Kinship verification loss:* Intuitively, given two batch-size samples $\mathbf{Q} = \{\mathbf{q}^1, \mathbf{q}^2, \cdots, \mathbf{q}^N\}$ and $\mathbf{T} = \{\mathbf{t}^1, \mathbf{t}^2, \cdots, \mathbf{t}^N\}$, the main loss function is formulated as:

$$L_{NCE} = -\frac{1}{N} \sum_{i=1}^{N} \log \frac{\exp(\cos(\mathbf{F}_\mathbf{q}^i, \mathbf{F}_\mathbf{t}^j)/\tau)}{\sum_{j=0}^{N} \exp(\cos(\mathbf{F}_\mathbf{q}^i, \mathbf{F}_\mathbf{t}^j)/\tau)} \quad (9)$$

where $N$ is the mini-batch size.

*2) De-aging feature learning loss:* For the de-aging feature learning, we first find the maximum of $\rho$ by freezing the backbone and residual factorization module, and train the canonical mapping module with three stacked FC-ReLU layers. Then, when fix C, we train the backbone and residual factorization module to reduce the correlation between $\mathbf{f}_{id}^i$ and $\mathbf{f}_{age}^i$. The objective loss function is

$$L_{de-aging}^q = -\frac{1}{N} \sum_{i=1}^{N} \min_{K,R} \max_C (|\rho(C(\mathbf{f}_{id}^i)), C(\mathbf{f}_{age}^i)|),$$
$$L_{de-aging}^t = -\frac{1}{N} \sum_{i=1}^{N} \min_{K,R} \max_C (|\rho(C(\mathbf{f}_{id}^i)), C(\mathbf{f}_{age}^i)|). \quad (10)$$

The derivative of $\rho$ with respect to $C(\mathbf{f}^i)$ is shown as follows:

$$-\frac{\delta\rho}{\delta C(\mathbf{f}_{id}^i)} = \sqrt{\frac{C(\mathbf{f}_{age}^i) - \mu_{age}^i}{\sigma_{id}^2 + \epsilon}\sqrt{\sigma_{age}^2 + \epsilon}} - \frac{(C(\mathbf{f}_{id}^i) - \mu_{id}) \cdot \rho}{\sigma_{id}^2 + \epsilon}$$
$$-\frac{\delta\rho}{\delta C(\mathbf{f}_{age}^i)} = \sqrt{\frac{C(\mathbf{f}_{id}^i) - \mu_{id}^i}{\sigma_{id}^2 + \epsilon}\sqrt{\sigma_{age}^2 + \epsilon}} - \frac{(C(\mathbf{f}_{age}^i) - \mu_{age}^i) \cdot \rho}{\sigma_{age}^2 + \epsilon}, \quad (11)$$

where $\mu^i$ is mean of $C(\mathbf{f}^i)$, $\sigma^2$ is the variance of $C(\mathbf{f}^i)$, and $E$ is the constant parameter. $C(\mathbf{f}^i)$ denotes similarity metric. $\mathbf{f}^i$ and $\mathbf{f}^i$ are the features derived from the backbone $\mathbf{K}$ of $\mathbf{Q}$ and $\mathbf{T}$, respectively.

For supervising the learning of $\mathbf{f}_{id}^i$ and $\mathbf{f}_{id}^i$ we use a softmax cross-entropy loss to introduce much strict constraints, so that the age-invariant information can be decomposed well. It can be written as follows:

$$L_{id}^q = -\frac{1}{N} \sum_{i=1}^{N} \log \frac{\exp C(\mathbf{f}_{id}^i)}{\sum_{i=1}^{N} \exp C(\mathbf{f}_{id}^i)},$$
$$L_{id}^t = -\frac{1}{N} \sum_{i=1}^{N} \log \frac{\exp C(\mathbf{f}_{id}^i)}{\sum_{i=1}^{N} \exp C(\mathbf{f}_{id}^i)}. \quad (12)$$

Finally, we use the total loss to supervise the de-aging framework, and we have:

$$L^q = L_{de-aging}^q + L_{id}^q,$$
$$L^t = L_{de-aging}^t + L_{id}^t. \quad (13)$$

where $N$ is the mini-batch size.

*3) Race feature learning loss:* For race feature learning, the softmax cross-entropy loss is formulated as follows:

$$L_{race}^q = -\frac{1}{M} \sum_{i=1}^{M} \log \frac{\exp G(\mathbf{q}^i)}{\sum_{i=1}^{M} \exp G(\mathbf{q}^i)},$$
$$L_{race}^t = -\frac{1}{M} \sum_{j=1}^{M} \log \frac{\exp G(\mathbf{t}^j)}{\sum_{j=1}^{M} \exp G(\mathbf{t}^j)}, \quad (14)$$

where $M$ is the mini-batch size.

## IV. EXPERIMENTS AND RESULTS

In this section, we conduct extensive experiments on some publicly-available datasets to evaluate the performance of different methods, and the effectiveness of the proposed method is validated by comparing some state-of-the-art algorithms.

### A. Datasets

There are many commonly-used datasets for kinship analysis, *i.e.,* **KinFaceW-I** [11], **KinFaceW-II** [11], **UBKinFace** [46] and **TSKinFace** [47]. Since these datasets are based on web crawling technology, they are easily interfered by the real-world complicated environment. As such, it is important but challenging to train these datasets. Here, we choose **CACD** [44] that has the similar environmental noise with kinship datasets as our de-aging dataset to enhance the robustness of the learning networks. The detailed description of each dataset is illustrated as follows and shown in Table II.

**CACD** [44] is a large-scale dataset with 163,446 images collected from 2000 celebrities ranging in age from 16 to 62 by web crawler technology. The dataset also provides detailed information on key points of 16 faces.

**KinFaceW-I** [11] has four typical types of kin relations: Father-Son (F-S) 156 pairs, Father-Daughter (F-D) 134 pairs, Mother-Son (M-S) 116 pairs and Mother-Daughter (M-D) 127 pairs, respectively. This dataset contains 1066 unconstrained face images from 533 people.

**KinFaceW-II** [11] has four representative types of kin relations: Father-Son (F-S), Father-Daughter (F-D), Mother-Son (M-S), and Mother-Daughter (M-D), respectively. Each type contains 250 unconstrained face image pairs selected from 1000 people. The difference between KinFaceW-II [11] and KinFaceW-I [11] is that the image of each parent-offspring pair from KinFaceW-II is collected from the same photograph.

**UBKinFace** [46] involves multiple age groups from young children, their young parents to old parents, which contains 200 triplets. It was collected more than 1,000 images from public figures (celebrities and politicians).

**TSKinFace** [47] includes 2589 people collected from publicly available figures (celebrities and politicians), which has no restrictions in terms of pose, lighting, expression, background, race, image quality, etc.

### B. Baselines

Some popular supervised learning based algorithms are applied to validate the effectiveness of different datasets under the same experimental configurations [11], such that the experimental results are reliable and convincing. We also compare our unsupervised method with these supervised algorithms. We simply list the description on each algorithm as follows:

*1) Shallow learning-based models for kinship verification:*

- **ASML** [15]: This method employs an adversarial metric learning to build a similarity metric.
- **LDA, MFA, WGEML** [48]: In this paper, we use multiple conventional feature learning models, such as LDA (Linear Discriminant Analysis), MFA (Marginal Fisher Analysis) and WGEML (Weighted Graph Embedding Based Metric Learning), to get multiple similarity metrics.
- **DMML** [4]: This method uses multiple features derived from different descriptors to learn multiple distance metrics.
- **NRML** [11]: The method uses NRML (Neighborhood Repulsed Metric Learning) to determine a distance metric.
- **L$M^3$L** [14]: This method uses multiple feature descriptors to extract various features for kinship verification.
- **DDMML** [12]: This method proposes a discriminative deep multi-metric learning method to maximize the correlation of different features for each sample.
- **MKSM** [16]: This method uses a multiple kernel similarity metric (MKSM) to combine multiple basic similarities for the feature fusion.
- **KINMIX** [49]: This method verifies kin relations by using a KinMix method to generate positive samples for data augmentation.

*2) Deep learning-based models for kinship verification:*

- **CNN-Basic, CNN-Points** [5]: For this method, we use the deep CNN model to extract features and classify the kin relation.
- **SMCNN** [24]: This method uses the similarity metric based CNNs to verify kin relation.
- **DTL** [26]: This method uses a transfer learning strategy and triangular similarity metric to train model, and leverages both face and the mirror face to increase robustness and accuracy.
- **CFT** [25]: This method combines transfer learning-based CNN and metric learning (NRML or other metrics) to get the final features.
- **DKV** [6]: This method uses the LBP features as the first input of an auto-encoder network, and then uses a metric learning for prediction.
- **AdvKin, E-AdvKin** [50]: This method uses an adversarial convolutional network with residual connections for facial kinship verification.
- **GKR** [51]: This method employs a graph-based kinship reasoning (GKR) network for kinship verification.

*Remark 1*: Although fcDBN [27] is one of the state-of-the-art methods, it is based on the Deep Belief Network (DBN) and even conducts a number of additional quantitative analyses of human performance on kinship verification. In addition, DBN needs greedy layer-wise training which is complicated and unstable for large-scale learning. However, our algorithm and mentioned baselines are based on CNN models without any additional observations. Therefore, it is unfair to compare it with our algorithm and mentioned baselines. Considering different backbones and preprocessing methods, we decided not to compare the results with this algorithm.

### C. Experimental Settings

*1) De-aging training:* The backbone consists of 4 stages with respectively 3, 4, 10, 3 stacked residual blocks. Each

block has three stacked units of 3*3. Finally, a FC layer outputs 512 channels. Moreover, for performance validation, we use MTCNN [52] to detect and align the face region and only conduct central cropping into 112 by 112. All experiments train on **CACD** [44]. Meanwhile, in an adversarial loop, the training run the canonical correlation maximizing process for 20 iterations, then change to feature correlation minimizing process for 50 iterations referring to [42]. Furthermore, we train the de-aging network by utilizing SGD(stochastic gradient descent) with a batch size 64 and a initial learning rate of 0.0005 and decrease it to 0.001 after the second epoch. The momentum is 0.9.

We first train the Canonical Mapping Module to find the maximum of $\rho$ by freezing Backbone, Residual Factorization Module, and optimize Canonical Mapping Module with SGD(stochastic gradient descent). Then we fix Canonical Mapping Module, at the same time train Backbone, Residual Factorization Module to reduce the correlation between $\mathbf{f}_{id}$ and $\mathbf{f}_{age}^l$ with SGD (stochastic gradient descent).

*2) Kinship verification:* To validate the performance of different methods for kinship verification, all experiments are performed on the **KinFaceW-I** [11], **KinFaceW-II** [11], **UBKinFace** [46] and **TSKinFace** [47] datasets. As shown in Table II, we use a generic framework to extract patch features, and each backbone is composed of a 30-block CNN. It consists of 3 stages with 10, 10, 10 stacked residual blocks, respectively. Each residual block has 3 Conv units respectively 1*1, 3*3, 1*1. Finally, after feature fusion, a linear layer outputs the initial face features with 1024 dimensionality. We train the DCML network by utilizing SGD (stochastic gradient descent) with a batch size 128. The momentum coefficient in momentum update $m$ is 0.999. The initial learning rate is set to 0.0001 and will decrease to 0.001 after the second epoch. The momentum is set to 0.9. Following the existing works [5], [11], [17], due to the different learning strategy, we only perform five-fold cross-validation on all datasets, and all face images are aligned and centrally cropped into 112 x 112 for the de-aging model, 64 x 64 for DCML framework, and 224 x 224 for race extraction, respectively. The memory bank size is 65536. According to the previous works [5], [11], [17], for supervised learning, each fold contains positive samples (with kinship relation) and negative samples (without kinship relation) with the same ratio, and the images in all relationships are roughly equal in all folds. Notably, for our unsupervised learning, the ratio of positive samples and negative samples is $1:N$, where $N$ is the mini-batch size. More importantly, all experiments follow 80%–20% protocol, *i.e.,* 80% sample images for training and the remained 20% samples for testing.

For multi-modal feature fusion, we apply simply concatenation operation, and the adaptive feature fusion mechanism is used to change the fusion weights according to the importance of features automatically. Finally, we verify our method by a linear classifier. The self-supervised learning is operated on four widely-used kinship datasets following a linear classifier (three fully-connected layers). Moreover, we train the classifier by the output of the adaptive average pooling layer. Meanwhile, we employ Top-1 and Top-5 classification accuracy as our evaluation accuracy. Particularly, the related hyper-parameter $r$ and extensive experiments will be discussed in Sec F: Ablation Study.

TABLE II: Family-based characteristics in experiments.

| Dataset | No. family | No. people | No. samples | Kin relations | Multiple images |
|---|---|---|---|---|---|
| TSKinFace [47] | – | 2,589 | 787 | 4 | No |
| KinFaceW-I [11] | – | 533 | 1,066 | 4 | No |
| KinFaceW-II [11] | – | 1,000 | 2,000 | 4 | No |
| UB KinFace [46] | – | 400 | 600 | 4 | No |

TABLE III: Backbone Structure based on ResNet.

| Output size | Residual block |
|---|---|
| 24 ×24 | conv, 7 ×7, 64, *stride*1<br>max pool, 3 ×3, *stride*2 |
| 24 ×24 | $\begin{bmatrix} conv & 1 \times 1, & 16 \\ conv & 3 \times 3, & 16 \\ conv & 1 \times 1, & 64 \end{bmatrix} \times 10$ |
| 12 ×12 | $\begin{bmatrix} conv & 1 \times 1, & 32 \\ conv & 3 \times 3, & 32 \\ conv & 1 \times 1, & 128 \end{bmatrix} \times 10$ |
| 6 ×6 | $\begin{bmatrix} conv & 1 \times 1, & 64 \\ conv & 3 \times 3, & 64 \\ conv & 1 \times 1, & 256 \end{bmatrix} \times 10$ |
| 1 ×1 | global average pool, fc |

***Remark 2***: It should be noted that our work is an ***unsupervised*** kinship verification framework, which is different from supervised ones. In our experiments, we found that there were very limited unsupervised kinship verification algorithms. Therefore, we compare our unsupervised results with these supervised algorithms, the results of which are directly cited from the original papers.

*3) Race training:* To effectively extract race feature, we employ **Resnet-50** [43] as our backbone. The network is pre-trained on **ImageNet** [28] dataset with freezed weights and bias. We use **CACD** [44] as our basic race dataset to finetune Resnet-50. We train the race model by utilizing Adam optimizer with an initial learning rate of 0.0001 and batch size 64. The momentum is set to 0.9. The learning rate is decreased by a factor of 10 after the second epoch.

*D. Evaluation Metrics*

To make an intuitive comparison of our method and other algorithms, we evaluate our kinship verification with the state-of-the-art algorithms on the Mean Verification Accuracy score. It can be defined as follows:

$$ACC = \frac{TP + TN}{P + N} * 100\% = \frac{TP}{P} * 100\%, \quad (15)$$

where $N = 0$ and $TN = 0$. $TP$ means the top-k prediction is $P$, and the true value is $P$. $TN$ means both prediction and true values are $N$. $P + N$ is the total training samples, which is the mini-batch size in our training process.

*E. Results and Analysis*

We use four widely-used kinship datasets to verify our model shown in Table IV. Table V shows our unsupervised learning results by comparing some state-of-the-art supervised methods. Obviously, the performance of the proposed unsupervised method can be comparable to these popular supervised kinship verification methods. In comparison with previous supervised results, the unsupervised method could be even superior to some state-of-the-art kinship verification methods.

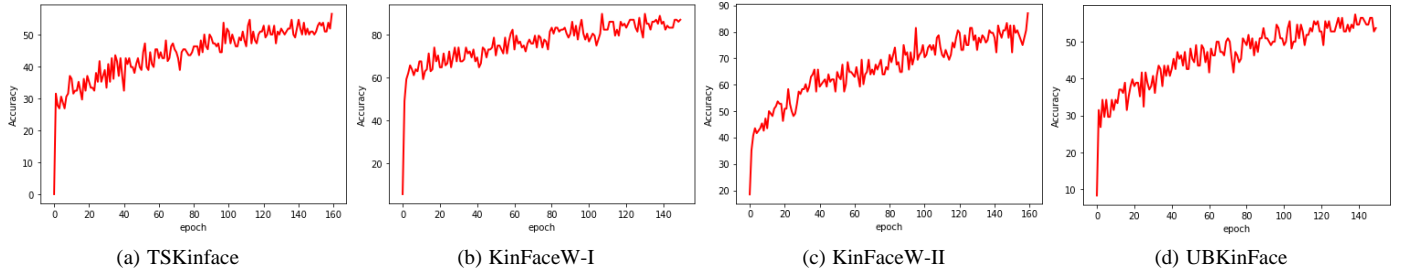

Fig. 3: The Top-5 testing accuracy curves of the proposed method. From left to right, these figures denote the results on the TSKinface dataset, KinFaceW-I dataset, KinFaceW-II dataset and UBKinFace dataset, respectively.

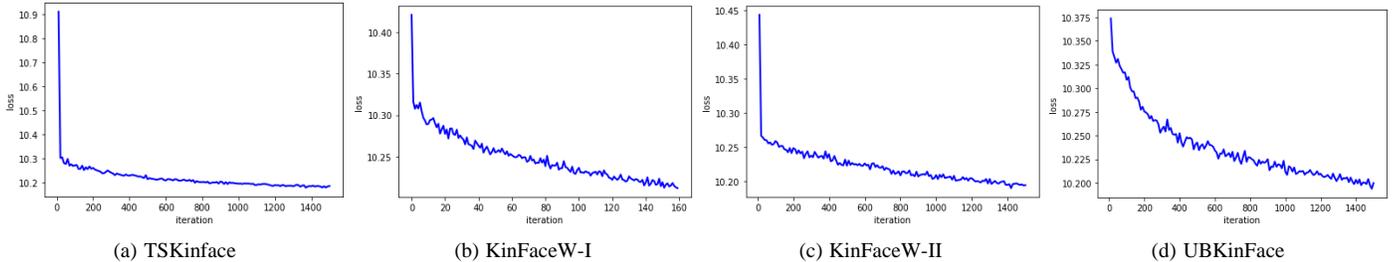

Fig. 4: The training loss curves of proposed method. From left to right, these figures denote the results on the TSKinface dataset, KinFaceW-I dataset, KinFaceW-II dataset and UBKinFace dataset, respectively.

TABLE IV: The TOP-k accuracies (%) of the proposed DCML model on different datasets for kinship verification.

| Dataset | F-S | F-D | M-S | M-D | Avg |
|---|---|---|---|---|---|
| TOP-1 | | | | | |
| TSKinFace [47] | 35.8 | 31.6 | 30.8 | 31.2 | 32.4 |
| KinFaceW-I [11] | 28.5 | 37.9 | 30.8 | 26.4 | 30.9 |
| KinFaceW-II [11] | 35.4 | 30.9 | 36.7 | 32.8 | 34 |
| UB KinFace [46] | 24.1 | 25.5 | 24.9 | 22.9 | 24.4 |
| TOP-5 | | | | | |
| TSKinFace [47] | 71.4 | 78.7 | 69.6 | 73.4 | 73.3 |
| KinFaceW-I [11] | 78.2 | 80.8 | 81.4 | 78.2 | 79.7 |
| KinFaceW-II [11] | 82.7 | 81.6 | 77.1 | 82.3 | 80.9 |
| UB KinFace [46] | 65.8 | 66.9 | 67.8 | 70.5 | 67.8 |

The superior performances of our unsupervised method may benefit from the following reasons. First, our multi-modal learning can capture the intrinsic underlying structure and uncover their implicit complementary advantages so that different modalities features can be treated as composite features to represent a deeper knowledge and share rich semantics. Second, we construct an effective adaptive multi-modal feature fusion mechanism, which can decrease the information redundancy and complexity between channels. Meanwhile, such mechanism can focus on the most informative components of feature maps to strengthen the dependencies between different properties. Third, we employ stronger self-supervised learning to explore deeper the latent information embedded in the raw interior structure of data. As such, it can generate self-learned semantic information. Hence, we could further improve the distinguishability of the learned features and mitigate the implicit variations from the original facial images.

Additionally, the proposed unsupervised method gets better improvements benefiting from the following training reasons. First, we use a multi-view strategy to represent the distinct sets of features under the same set of the underlying objects, which can capture more comprehensive and related information from multiple perspectives. Notably, most previous works are not focused on multi-view images, leading to missing view data. This missing view data results in the lack of facial details. Moreover, kinship datasets are based on web crawler technology. They have a lot of environmental noise. Hence, employing Multi-view learning can be applied to reduce the noise and learn more refined facial details. Second, we leverage face attributes as multi-modal features to obtain more natural and powerful discriminative information. Particularly, the facial details can be changed by aging. And the faces under the similar two age periods have a similar texture distribution. Meanwhile, the similarity degree of the face in the two age classes is almost inversely proportional to their age deviation. Therefore, the age-invariant features can represent more natural facial details, and age-related transformations are more significant for face recognition. Figures 3 and 4 show training loss and test accuracy.

TABLE V: Comparison results of different methods on different datasets for kinship verification.

| Learning strategy | Methods | TSKinFace | | | | | KinFaceW-I | | | | | KinFaceW-II | | | | | UB KinFace | | | | |
|---|---|---|---|---|---|---|---|---|---|---|---|---|---|---|---|---|---|---|---|---|---|
| | | F-S | F-D | M-S | M-D | Avg | F-S | F-D | M-S | M-D | Avg | F-S | F-D | M-S | M-D | Avg | F-S | F-D | M-S | M-D | Avg |
| Supervised Learning | ASML [15] | – | – | – | – | – | 82.7 | 76.1 | 78.0 | 81.6 | 79.6 | 85.8 | 78.0 | 81.8 | 78.6 | 81.1 | – | – | – | – | – |
| | MLDA [48] | – | – | – | – | – | 76.6 | 71.2 | 77.7 | 76.4 | 75.5 | 86.6 | 74.4 | 81.0 | 78.8 | 80.2 | – | – | – | – | – |
| | MMFA [48] | – | – | – | – | – | 77.9 | 72.0 | 75.2 | 75.6 | 75.2 | 85.6 | 73.2 | 80.4 | 77.2 | 79.1 | – | – | – | – | – |
| | NRML [11] | – | – | – | – | 74.2 [a] | 76.3 | 69.8 | 77.2 | 75.9 | 74.8 | 82.6 | 68.6 | 76.6 | 73.0 | 75.2 | – | – | – | – | – |
| | WGEML [48] | – | – | – | – | – | 78.5 | 73.9 | 80.6 | 81.9 | 78.7 | 88.6 | 77.4 | 83.4 | 81.6 | 82.8 | – | – | – | – | – |
| | DMML [4] | – | – | – | – | – | 74.5 | 69.5 | 69.5 | 75.5 | 72.3 [b] | 78.5 | 76.5 | 78.5 | 79.5 | 78.3 | – | – | – | – | 72.25 |
| | DDMML [12] | – | – | – | – | – | 86.4 | 79.1 | 81.4 | 87.0 | 83.5 | 87.4 | 83.8 | 83.2 | 83.0 | 84.3 | – | – | – | – | – |
| | L$M^3$L [14] | – | – | – | – | – | – | – | – | – | – | 82.4 | 74.2 | 79.6 | 78.7 | 78.7 | – | – | – | – | – |
| | MKSM [16] | – | – | – | – | – | 83.65 | 81.35 | 79.69 | 81.16 | 81.46 | 83.80 | 81.20 | 82.40 | 82.40 | 82.45 | – | – | – | – | – |
| | KINMIX [49] | – | – | – | – | – | 75.6 | 76.5 | 78.5 | 83.5 | 78.5 | 89.6 | 87.2 | 91.2 | 90.6 | 89.7 | – | – | – | – | – |
| | NUAA [47] | – | – | – | – | – | 86.25 | 80.64 | 81.03 | 83.93 | 82.96 | 84.40 | 81.60 | 82.80 | 81.60 | 82.50 | – | – | – | – | – |
| | M-NUAA [47] | – | – | – | – | – | 87.84 | 85.47 | 86.16 | 87.50 | 86.74 | 88.40 | 86.20 | 86.00 | 85.20 | 86.45 | – | – | – | – | – |
| | CNN-Basic [5] | – | – | – | – | – | 75.7 | 70.8 | 73.4 | 79.4 | 74.83 | 84.9 | 74.8 | 88.3 | 88.5 | 85.33 | – | – | – | – | – |
| | CNN-Points [5] | – | – | – | – | – | 76.1 | 71.8 | 78.0 | 84.1 | 77.5 | 89.4 | 81.9 | 89.9.0 | 92.4 | 88.4 | – | – | – | – | – |
| | SMCNN [24] | – | – | – | – | – | 75.0 | 75.0 | 68.7 | 72.2 | 72.73 | 75.0 | 79.0 | 78.0 | 85.0 | 79.25 | – | – | – | – | – |
| | DTL [26] | – | – | – | – | – | 86.45 | 89.62 | 89.57 | 88.8 | 88.61 | 83.4 | 89.4 | 87.0 | 87.0 | 86.7 | – | – | – | – | – |
| | CFT [25] | – | – | – | – | – | 78.8 | 71.7 | 77.2 | 81.9 | 77.4 | – | – | – | 65.5 | – | 77.4 | 76.6 | 79.0 | 83.8 | 79.3 |
| | DKV [6] | – | – | – | – | – | 71.8 | 62.7 | 66.4 | 66.6 | 66.9 | 73.4 | 68.2 | 71.0 | 72.8 | 71.3 | – | – | – | – | – |
| | AdvKin [50] | – | – | – | – | – | 75.7 | 78.3 | 77.6 | 83.1 | 78.7 | 88.4 | 85.8 | 88.0 | 89.8 | 88.0 | – | – | – | – | 81.4 |
| | E-AdvKin [50] | – | – | – | – | – | 76.6 | 78.4 | 78.4 | 86.2 | 79.6 | 91.6 | 85.2 | 90.2 | 92.4 | 89.9 | – | – | – | – | 80.4 |
| | GKR [51] | – | – | – | – | – | 79.5 | 73.2 | 78.0 | 86.2 | 79.2 | 90.8 | 86.0 | 91.2 | 94.4 | 90.6 | – | – | – | – | – |
| Supervised Learning | DCML | 89.2 | 87.5 | 88.4 | 89.1 | **88.6** | 92.2 | 93.2 | 93.5 | 93.0 | **93.0** | 92.8 | 93.0 | 94.3 | 93.8 | **93.5** | 88.4 | 89.3 | 90.7 | 91.2 | **89.9** |
| Unsupervised Learning | DCML(Top-5) | 71.4 | 78.7 | 69.6 | 73.4 | **73.3** | 78.2 | 80.8 | 81.4 | 78.2 | **79.7** | 82.7 | 81.6 | 77.1 | 82.3 | **80.9** | 65.8 | 66.9 | 67.8 | 70.5 | **67.8** |

a: Red is the best performance in the previous works.
b: Blue is the worst performance in the previous works.

TABLE VI: Comparison results (top-K accuracy) of different reduction ratios $r_1$ & $r_2$ on all datasets for kinship verification.

| | | TOP-1 | | | | TOP-5 | | | |
|---|---|---|---|---|---|---|---|---|---|
| Ratio $r_1$ | Ratio $r_2$ | TSKinFace [47] | KinFaceW-I [11] | KinFaceW-II [11] | UB KinFace [46] | TSKinFace [47] | KinFaceW-I [11] | KinFaceW-II [11] | UB KinFace [46] |
| 2 | 2 | 31.8 | 29.6 | 33.8 | 23.2 | 73.1 | 76.7 | 77.6 | 65.4 |
| | 4 | 32.1 | 30.1 | 34.1 | 23.5 | 73.2 | 76.9 | 77.9 | 65.2 |
| | 8 | 31.4 | 30.9 | 32.7 | 23.8 | 72.7 | 76.6 | 78.5 | 68.5 |
| | 16 | 32.1 | 28.7 | 33.2 | 22.9 | 70.8 | 75.7 | 73.4 | 65.5 |
| 4 | **2** | **32.4** | **30.9** | **34** | **24.4** | **73.3** | **79.7** | **80.9** | **67.8** |
| | 4 | 32.3 | 30.7 | 33.5 | 24.1 | 73.4 | 77.1 | 79.1 | 67.9 |
| | 8 | 31.9 | 30.4 | 32.9 | 23.7 | 72.5 | 77.3 | 79.5 | 67.1 |
| | 16 | 32.0 | 29.5 | 33.6 | 22.1 | 71.8 | 77.6 | 78.5 | 67.5 |
| 8 | 2 | 31.7 | 30.8 | 33.6 | 23.1 | 72.1 | 75.9 | 76.9 | 66.4 |
| | 4 | 31.4 | 28.9 | 32.9 | 25.1 | 73.3 | 76.4 | 78.5 | 67.8 |
| | 8 | 32.2 | 30.6 | 33.9 | 24.3 | 72.7 | 76.1 | 77.2 | 65.5 |
| | 16 | 31.5 | 20.7 | 34.2 | 23.2 | 72.6 | 75.9 | 77.4 | 67.5 |
| 16 | 2 | 30.6 | 29.6 | 30.6 | 23.1 | 72.4 | 76.1 | 77.1 | 64.5 |
| | 4 | 29.6 | 29.9 | 31.5 | 22.4 | 73.4 | 76.2 | 76.5 | 66.7 |
| | 8 | 31.7 | 30.7 | 27.9 | 23.8 | 72.1 | 76.9 | 78.3 | 65.8 |
| | 16 | 32.1 | 30.3 | 27.1 | 22.7 | 71.5 | 77.7 | 78.4 | 67.2 |

TABLE VII: Comparison results of different modality combination for kinship verification on different datasets under $r_1 = 4$ and $r_2 = 2$.

| | | TOP-1 | | | | | TOP-5 | | | | |
|---|---|---|---|---|---|---|---|---|---|---|---|
| Dataset | Modality Combination | F-S | F-D | M-S | M-D | Avg | F-S | F-D | M-S | M-D | Avg |
| TSKinFace | face | 31.8 | 32.1 | 28.7 | 33.8 | 31.6 | 71.3 | 74.2 | 66.1 | 73.6 | 71.3 |
| | face + Race | 31.9 | 30.9 | 31.7 | 32.0 | 31.6 | 76.4 | 76.9 | 65.9 | 69.3 | 72.1 |
| | face + De-aging | 30.4 | 31.3 | 29.6 | 30.6 | 30.5 | 75.9 | 73.5 | 68.7 | 63.9 | 70.5 |
| | face + Race + De-aging | 35.8 | 31.6 | 30.8 | 31.2 | **32.4** | 71.4 | 78.7 | 69.6 | 73.4 | **73.3** |
| KinFaceW-I | face | 26.5 | 36.3 | 28.2 | 26.9 | 29.5 | 73.5 | 73.2 | 75.9 | 67.1 | 72.4 |
| | face + Race | 27.2 | 34.6 | 28.3 | 29.1 | 29.8 | 74.6 | 77.1 | 77.5 | 79.8 | 77.3 |
| | face + De-aging | 24.8 | 26.6 | 31.4 | 31.3 | 28.5 | 76.6 | 75.9 | 70.6 | 71.4 | 73.6 |
| | face + Race + De-aging | 28.5 | 37.9 | 30.8 | 26.4 | **30.9** | 78.2 | 80.8 | 81.4 | 78.2 | **79.7** |
| KinFaceW-II | face | 31.2 | 33.7 | 29.6 | 25.2 | 30.0 | 78.1 | 75.6 | 77.9 | 76.2 | 77.0 |
| | face + Race | 32.2 | 30.8 | 34.4 | 33.1 | 32.6 | 80.6 | 79.7 | 76.1 | 79.3 | 79.0 |
| | face + De-aging | 32.6 | 30.1 | 29.5 | 33.1 | 31.3 | 81.1 | 76.8 | 82.6 | 77.9 | 79.6 |
| | face + Race + De-aging | 35.4 | 30.9 | 36.7 | 32.8 | **34** | 82.7 | 81.6 | 77.1 | 82.3 | **80.9** |
| UB | face | 22.0 | 21.5 | 19.9 | 21.4 | 21.2 | 62.7 | 60.8 | 61.9 | 63.5 | 62.2 |
| | face + Race | 22.7 | 23.6 | 24.1 | 21.8 | 23.1 | 64.2 | 65.1 | 67.5 | 67.8 | 66.2 |
| | face + De-aging | 23.7 | 22.6 | 25.9 | 24.1 | 24.1 | 60.6 | 69.8 | 64.6 | 65.9 | 65.2 |
| | face + Race + De-aging | 24.1 | 25.5 | 24.9 | 22.9 | **24.4** | 65.8 | 66.9 | 67.8 | 70.5 | **67.8** |

## F. Ablation Study

In this section, we conduct extensive ablation studies to verify the indispensability of different components in our DCML framework. Typically, we experiment on different $r$ combination, different property combination and weights selection. Moreover, we also explore our deep collaborative multi-modal learning on supervised learning extension.

TABLE VIII: TOP-1 Performance Comparisons With Other Deep Multi-modal Feature Fusion Learning Networks.

| Methods | TSKinFace | | | | | KinFaceW-I | | | | | KinFaceW-II | | | | | UB KinFace | | | | |
|---|---|---|---|---|---|---|---|---|---|---|---|---|---|---|---|---|---|---|---|---|
| | F-S | F-D | M-S | M-D | Avg | F-S | F-D | M-S | M-D | Avg | F-S | F-D | M-S | M-D | Avg | F-S | F-D | M-S | M-D | Avg |
| Multi-abstract Fusion [53] | 26.6 | 28.4 | 23.2 | 22.1 | 25.1 | 23.5 | 27.3 | 29.0 | 22.9 | 25.7 | 29.2 | 31.3 | 32.5 | 33.2 | 31.6 | 21.1 | 24.9 | 20.5 | 17.3 | 21 |
| Low-rank-Multimodal-Fusion [54] | 36.1 | 32.8 | 28.0 | 34.1 | **32.8** | 29.4 | 31.9 | 29.2 | 32.4 | 30.7 | 33.5 | 31.1 | 32.6 | 34.5 | 32.9 | 25.0 | 24.7 | 28.7 | 22.2 | **25.2** |
| SPP [55] | 21.0 | 20.7 | 17.7 | 20.2 | 19.9 | 23.3 | 22.1 | 21.0 | 19.1 | 21.4 | 24.7 | 23.9 | 24.5 | 23.2 | 24.1 | 15.7 | 16.1 | 18.4 | 19.3 | 17.4 |
| MDLN [56] | 26.4 | 22.6 | 21.3 | 20.8 | 22.8 | 23.4 | 19.8 | 26.7 | 20.4 | 22.6 | 22.9 | 25.7 | 24.8 | 21.9 | 23.8 | 18.4 | 20.7 | 19.3 | 19.9 | 19.6 |
| Manually-assigned Weights (Top-1) | 28.4 | 29.3 | 20.7 | 21.2 | 24.9 | 22.2 | 28.7 | 23.5 | 23 | 24.4 | 23.8 | 27.9 | 30.4 | 26.7 | 27.2 | 24.1 | 25.5 | 24.9 | 22.9 | 24.4 |
| DCML(Top-1) | 35.8 | 31.6 | 30.8 | 31.2 | 32.4 | 28.5 | 37.9 | 30.8 | 26.4 | **30.9** | 35.4 | 30.9 | 36.7 | 32.8 | **34** | 24.1 | 25.5 | 24.9 | 22.9 | 24.4 |

TABLE IX: TOP-5 Performance Comparisons With Other Deep Multi-modal Feature Fusion Learning Networks.

| Methods | TSKinFace | | | | | KinFaceW-I | | | | | KinFaceW-II | | | | | UB KinFace | | | | |
|---|---|---|---|---|---|---|---|---|---|---|---|---|---|---|---|---|---|---|---|---|
| | F-S | F-D | M-S | M-D | Avg | F-S | F-D | M-S | M-D | Avg | F-S | F-D | M-S | M-D | Avg | F-S | F-D | M-S | M-D | Avg |
| Multi-abstract Fusion [53] | 68.6 | 64.3 | 67.4 | 64.0 | 66.1 | 65.3 | 66.5 | 67.2 | 63.2 | 65.6 | 69.8 | 66.7 | 69.5 | 65.2 | 67.8 | 58 | 62.3 | 48.6 | 62.2 | 57.8 |
| Low-rank-Multimodal-Fusion [54] | 70.5 | 71.5 | 75.4 | 73.2 | 72.7 | 78.1 | 74.3 | 76.5 | 79.4 | 77.1 | 83.2 | 80.6 | 78.9 | 79.4 | 80.5 | 67.4 | 62.9 | 64.6 | 67.2 | 65.5 |
| SPP [55] | 55.3 | 58.5 | 45.8 | 53.1 | 58.2 | 62.1 | 63.3 | 66.0 | 60.4 | 63 | 70.5 | 68.4 | 67.3 | 70.1 | 69.1 | 60.1 | 58.8 | 55.4 | 57.3 | 57.9 |
| MDLN [56] | 66.4 | 63.1 | 60.8 | 63.2 | 63.4 | 67.2 | 65.6 | 62.5 | 66.0 | 65.3 | 69.2 | 68.8 | 65.9 | 66.1 | 67.5 | 53.1 | 61.2 | 54.6 | 57.2 | 56.5 |
| Manually-assigned Weights (Top-5) | 62.7 | 66.1 | 58 | 67.6 | 63.6 | 71.3 | 76.4 | 72 | 75.5 | 73.8 | 76.3 | 75.9 | 72.3 | 76.4 | 75.2 | 65.6 | 62.1 | 65.2 | 62.3 | 63.8 |
| DCML(Top-5) | 71.4 | 78.7 | 69.6 | 73.4 | **73.3** | 78.2 | 80.8 | 81.4 | 78.2 | **79.7** | 82.7 | 81.6 | 77.1 | 82.3 | **80.9** | 65.8 | 66.9 | 67.8 | 70.5 | **67.8** |

*1) Reduction ratio r:* We evaluate the reduction ratio $r_1$ and $r_2$ introduced in Eq. (2) and Eq. (7), respectively. Under the effect of this hyper-parameter, we perform experiments based on the proposed backbone with a range of different $r$ combinations shown in Table VI. In these experiments, we do not consider the computational cost, but only concern the performance. The comparison results shown Table VI demonstrates that the ratio combination is not consistent under the best results in different datasets. We choose a better combination to trade-off the variations cross different datasets. These experimental results also verify that the performance is relatively stable *w.r.t.* a range of reduction ratio combinations. Notably, using the same ratios may be not the best combination choice. Moreover, increasing or decreasing the reduction ratio does not greatly influence the performance, and the performance is not monotonic when changing $r$ changing. In our experiments, we set $r_1 = 4$ and $r_2 = 2$ as the best combination.

*2) Property combination:* In this paper, we propose a novel deep collaborative multi-modal learning (DCML) to enhance the representation capability of the learned features, which can aggregate multiple visual knowledge for unsupervised kinship verification. The experimental results mentioned above have shown the advantages of the proposed unsupervised method. Here, we estimate the performance of different modality combinations by systematically changing the feature fusion strategy. Especially, the proposed multi-modal feature fusion framework consists of three kinds of modalities, *i.e.,* the original facial image visual features, the de-aging features, and the race features. Table VII shows the experimental results *w.r.t.,* different modality combinations. In comparison to a single modality, combining some meaningful visual properties, such as race and age, can further improve the representation capabilities of the learned features. From the experimental results, we can observe that the proposed DCML further validates the indispensability and effectiveness of different modalities combinations. The multi-modal strategy will take advantage of their implicit complementary advantages to learn the significant details in faces.

*3) Weights selection:* How to select the best weighting strategy is one of the most crucial research topics in multi-modal learning. To testify the effectiveness of adaptive modality fusion, we compare with some multi-feature fusion components proposed in similar works *i.e.* Multi-abstract Fusion [53], Low-rank-Multimodal-Fusion [54], SPP [55], MDLN [56] to demonstrate the superiority of our adaptive feature fusion mechanism against the widely-used tricks, such as concatenation or manually-assigned weights or other similar operations. Here, we set the weight of each point as $\frac{1}{n}$, where $n$ is the number of the used modalities in multi-modal feature fusion module or used patches in facial image visual feature extraction module. The comparison results are summarized in Tables VIII and IX.

The qualitative and quantitative experiments demonstrate that our proposed adaptive feature fusion mechanism is always superior to concatenation or manually-assigned weighting strategies and some of multi-feature fusion method. Notably, multiple modalities can help learn more natural details and capture complementary properties embedded in multiple modalities. Moreover, our proposed feature fusion mechanism can mitigate the information redundancy effectively and simplify the overall complexity. It can focus on the most informative components adaptively, and the importance of different modalities is well considered in the feature learning process, leading to refined aggregation features instead of simple concatenation only.

*4) Deep collaborative multi-modal learning based on supervised learning:* To evaluate the effectiveness of our framework for *supervised* kinship verification, we further extend our deep collaborative multi-modal learning to a supervised learning diagram. Similar to the existing works [11], [48], we perform five-fold cross-validation on all datasets for model training. Meanwhile, all image sizes are the same as those in the unsupervised learning experiments. Notably, each parent matches the child randomly for negative samples, who are not the corresponding parent's real children. Moreover, each image of their parent-offspring pair only uses once in the negative samples. Following the widely-used splittings [4], [5], [12], all experiments for supervised learning algorithms are pre-trained on the FIW dataset [60] and follow the same 80% - 20% protocol, *i.e.,* the 80% sample images for training and the remained 20% for testing. That means all the experimental settings and evaluation metrics are the same as the previous unsupervised learning shown in Subsection IV-C.

TABLE X: Comparison results of different methods on FIW dataset for kinship verification.

| Methods | siblings | | | parent-child | | | | grandparent-grandchild | | | | Avg |
|---|---|---|---|---|---|---|---|---|---|---|---|---|
| | B-B | S-S | SIBS | F-D | F-S | M-D | M-S | GF-GD | GF-GS | GM-GD | GM-GS | |
| SphereFace [57] | 71.94 | 77.30 | 70.23 | 69.25 | 68.50 | 71.81 | 69.49 | 66.07 | 66.36 | 64.58 | 65.40 | 69.18 |
| VGG+DML [58] | – | – | 75.27 | 68.08 | 71.03 | 70.36 | 70.76 | 64.90 | 64.81 | 67.37 | 66.50 | 68.79 |
| ResNet+SDMLoss [59] | – | – | – | 69.02 | 68.60 | 72.28 | 69.59 | 65.89 | 65.12 | 66.41 | 64.90 | 69.47 |
| DCML(Top-1) | 46.6 | 51.4 | 49.8 | 47.5 | 41.6 | 49.4 | 48.8 | 37.2 | 25.4 | 31.9 | 32.4 | 42 |
| DCML(Top-5) | 79.2 | 73.0 | 64.5 | 76.3 | 73.2 | 77.6 | 70.5 | 53.7 | 57.6 | 52.8 | 55.4 | 66.7 |
| DCML(Supervised) | 90.5 | 87.6 | 81.9 | 91.7 | 89.3 | 90.2 | 89.7 | 73.6 | 79.8 | 76.3 | 70.5 | **83.7** |

Similarly, we also employ four widely-used kinship datasets to verify our proposed method. It is clear that the performance of the proposed method has achieved outstanding results. From the experiments shown in Table V, we intuitively find that our method represents stronger adaptability in face feature extraction. We can see that, compared to other popular algorithms, our framework has a big improvement by at least 4% on KinFaceW-I, while advances the performance at least 3% on KinFaceW-II. Moreover, we improve the performance on the UB Kin dataset for a total 9% gain on the averaged accuracy and a total 14% gain on the averaged accuracy on the TSKinFace dataset.

The proposed method gets clear improvements benefiting from followed reasons. First, previous methods did not consider the effect of multi-modal learning. The single modal learning can not satisfy the request that can capture correlations between different modalities. In particular, facial representation learning is a challenging task because it is strongly influenced by environmental conditions (illumination, age, and face expression). Hence, previous methods fail to capture the multiple underlying characteristics embedded in multiple modalities for effective kinship verification. Second, our adaptive feature fusion mechanism can select higher-level semantic features at the category-level prediction layer. This mechanism can pay more attention to simplify the complicated information and select more informative information to enhance the discriminability of the learned features. Third, supervised learning can get complementary information provided by the dataset and generate more high-level semantics to defense the large complex variations on face images, yielding state-of-the-art performance on different challenging benchmarks.

*5) On the large-scale dataset FIW:* Although extensive studies have been devoted to improving the robustness and discriminant of kinship verification systems, the related technology has not yet been suitable for real-world uses. In particular, current kinship systems are still facing several challenges *i.e.,* insufficient data, and more complex relations. These challenges lead to difficulty describing the actual data distributions of a family and capturing hidden factors affecting facial appearances. Therefore, a **Large-Scale Families in the Wild (FIW)** dataset [60] has been proposed for kinship learning. **FIW** is the largest dataset for kinship verification and includes approximately 656K face pairs of 10676 people, which are collected from 1000 different families. There are 11 different kin relations and enough for kinship-related tasks.

We employ the kinship verification evaluation on the FIW dataset in this sub-section to evaluate the proposed framework. Several state-of-the-art comparative methods are used to evaluate the dataset. Similar to the existing works [11], [48], we perform five-fold cross-validation for model training. Meanwhile, all experimental configurations are the same as those in the unsupervised learning and supervised learning experiments. Notably, each parent matches the child randomly for negative samples, who are not the corresponding parent's real children, and no family overlap between folds. All experiments follow the same training protocol and evaluation metrics shown in Subsection IV-C.

The comparison results reported in Table X have illustrated the feasibility and superiority of our DCML compared with some advanced kinship verification methods such as SphereFace [57], VGG+DML [58] and ResNet+SDMLoss [59]. Specifically, our proposed method improves the performance at least 13% in supervised learning. Moreover, we have achieved similar results in unsupervised learning compared to previous supervised-based works on FIW. From the results mentioned above, our proposed DCML can better represent the facial details to promote the distinguishable ability of the learned features and has verified our proposed model has a strong generalization by applying it on the large-scale dataset.

## V. CONCLUSIONS

In this paper, we proposed a novel deep collaborative multi-modal learning (DCML) method for unsupervised kinship verification, which jointly considers collective multi-modal learning, adaptive modality fusion mechanism, and self-supervised semantic enhancement. Notably, the proposed DCML method, for the first time, provided a new unsupervised learning framework for robust kinship estimation. Typically, our DCML takes advantage of the complementary correlations to cross multiple modalities in a self-adaptive interaction manner. Moreover, an adaptive feature fusion mechanism was designed to determine the importance of different modalities, which could flexibly build distinguishable knowledge and simplify the complicated information among channels. Besides, a self-supervised learning strategy was conceived to generate rich semantics. Meanwhile, the diversity of data was enriched to improve the discriminative abilities of the learned representation. Extensive experiments and analyses demonstrated the superb efficacy of the proposed method on unsupervised and supervised kinship analysis.

## VI. ACKNOWLEDGMENT

The authors would like to thank Hao Wang, Dihong Gong, Zhifeng Li, and Wei Liu for providing details of their de-aging framework and training methods to us, which greatly helps us to achieve the proposed method.